\documentclass[]{llncs}
\usepackage[T1]{fontenc}
\usepackage{graphicx}
\usepackage{tcolorbox}
\usepackage{xcolor}
\usepackage{float}
\usepackage{multirow}
\usepackage{diagbox}

\usepackage{booktabs} 
\usepackage{tabularx} 
\usepackage{amsmath} 
\usepackage{amssymb}

\AtBeginDocument{%
  }
    
\usepackage[utf8]{inputenc}
\usepackage{tcolorbox}
\tcbuselibrary{skins, raster}
\usepackage{xcolor}

\usepackage[utf8]{inputenc} 
\usepackage[T1]{fontenc}    
\usepackage{hyperref}       
\usepackage{url}            
\usepackage{booktabs}       
\usepackage{amsfonts}       
\usepackage{nicefrac}       
\usepackage{microtype}      
\usepackage{xcolor}         

\usepackage{cite}
\begin{document}
\title{English K\_Quantization of LLMs Does Not Disproportionately Diminish Multilingual Performance}
\titlerunning{Multilingual Quantization Performance Evaluation}
%
%
\author{Karl Audun Kagnes Borgersen\inst{1}\orcidID{0009-0004-8270-4411} \and
Morten~Goodwin\inst{1}\orcidID{0000-0001-6331-702X}}
\authorrunning{K.A. Borgersen et al.}

\institute{University of Agder, Grimstad, Norway\\
\email{karl.audun.borgersen@uia.no, morten.goodwin@uia.no}}
\maketitle              
\begin{abstract}
For consumer usage of locally deployed LLMs, the GGUF format and k\_quantization are invaluable tools for maintaining the performance of the original model while reducing it to sizes deployable with consumer-grade hardware. This size reduction is achieved by reducing the number of bits dedicated to each weight in the original model based on how important they are thought to be during model inference. A weight's importance is arrived at through the application of an 'importance matrix'—a relatively small text document meant to be representative of the LLM's standard use-cases. For the vast majority of quantized LLMs available online, this document is primarily written in English. It was therefore an open question whether performance on English language tasks was preserved through the sacrifice of multilingual performance and whether it can be preserved with alternate importance matrices. This article investigates these hypotheses by quantizing Llama3.3 70B on importance matrices written in three languages (English, Norwegian, and Malayalam) and evaluating them on the MixEval dataset in both English and Norwegian. All experiments yielded nonsignificant results, indicating that current quantization practices do not disproportionately harm multilingual performance.

\end{abstract}
\section{Introduction}
Large Language Models (LLMs) have had a remarkable impact on the daily lives of people over the last few years. Despite their global reach, these LLMs are generally formally trained to speak only a few widespread languages, such as English or Chinese. Considering the number of speakers for both English ($\sim1.5B$), and Mandarin Chinese ($\sim1.1B$), the choice to use these as a basis for an LLM's conversational ability is hardly surprising when compared to the $\sim5.4M$ total Norwegian speakers\cite{ethnologue200}. However, these LLMs often acquire an ability to speak these relatively uncommon languages regardless through their inclusion in an LLM's pre-training data\cite{languageRanker}. For example, while non-English languages were described as out-of-scope for the original llama2 release, its pre-training dataset consisted of $10.3\%$ non-English text of which $0.03\%$ was in Norwegian\cite{llama2}\footnote{This "incidental" acquisition of Norwegian language skills accounts for the vast majority of Norwegian-speaking LLMs today, for example, the 2042 models tagged as either Bokmål or Nynorsk on Huggingface's model repository\cite{huggingfacemodelsno}. There are a few prominent exceptions, however, such as the llama 3.2 finetunes released by the National Library of Norway\cite{nbllamapaper, nbllama} and the Nor-BERT family of LLMS\cite{norBERT}}. Until recently, only the extremely large proprietary LLMs of companies such as OpenAI or Anthropic could write Norwegian at a passable level, and this capability was fully beyond the available open-source\footnote{including open weights models with permissive licenses. Henceforth, these will both be referred to as "open-source"} models available. This changed with the release of newer open-source models such as version Llama3.3\cite{llama3herd} and Deepseek v3\cite{deepseekV3}. A qualitative sample of the difference between older and newer models is shown in Figure \ref{fig:nor-conv}.\\

The extreme multilingualism of modern LLMs has been achieved through many different training techniques, the most prominent of these being the sheer size of the models. The unique quirks of each language require the dedication of a significant number of weights, even if the exact nature and magnitude of these weights remains uncertain due to the black box nature of LLMs and Machine Learning (ML) as a whole\footnote{We dive deeper into this topic in Section \ref{sec:prev-inv}}. The amount of memory required to store and run inference on the weights of these models often means that running them is prohibitively expensive for most consumers to run locally. In an effort to combat this problem, researchers have released methods for quantizing ML models, i.e., to reduce the number of bits per weight while retaining as much of the original behavior as possible. As the information known by an LLM is innately tied to its weights, this quantization process will naturally remove some content from its knowledge base. It is uncertain, however, to what extent the multilingual nature of a model is harmed by this process. In particular, regarding the common methods of quantizing .gguf files that employ the use of an importance matrix, such as k\_quantization. These quantization methods, and those similar to them, do not apply their quantization uniformly across all the weights of a model; rather, they use an example text document to assess the importance of each weight and dedicate fewer bits to those deemed of lesser importance to the functioning of the model.\\
\colorlet{boxborder}{brown!60!black}   
\colorlet{boxbackground}{orange!5}     
\colorlet{boxtitle}{white!90!black}    
\colorlet{boxcontent}{black}           
\colorlet{notebg}{orange!10}            
\colorlet{norwegian}{teal!70!black}    
\colorlet{english}{red!50!black}       

\begin{figure*}[t]
    \begin{minipage}{\textwidth}
        \begin{tcolorbox}[
            title={Prompt:},
            colframe=boxborder,
            colback=boxbackground,
            coltitle=boxtitle,
            coltext=boxcontent,
            fonttitle=\bfseries,
            height=1.5cm,
            halign=center,
            valign=center
        ]
        Hvem var norges første konge?
        \end{tcolorbox}
        \begin{tcolorbox}[
            colframe=boxborder,
            colback=notebg,
            height=1cm,
            valign=center,
            nobeforeafter,
            boxrule=0.5pt,
            halign=center,
            valign=center
        ]
        \small Note: Prompt written in Norwegian Bokmål. Translation: "Who was the first king of Norway?"
        \end{tcolorbox}
        \vspace{2.0pt}
    \end{minipage}

    \begin{minipage}{0.48\textwidth}
        \begin{tcolorbox}[
            title={llama2 13B:},
            colframe=boxborder,
            colback=boxbackground,
            coltitle=boxtitle,
            coltext=boxcontent,
            fonttitle=\bfseries,
            height=3.5cm
        ]
        \color{norwegian}
        Hei, ja det er kjempefint å høre på deg! Du spør \color{english}after Norway's first king, ya? Well, let me tell ya, the
        first king of Norway was Harald Fairhair. He lived way...
        \end{tcolorbox}
        \begin{tcolorbox}[
            colframe=boxborder,
            colback=notebg,
            height=1.7cm,
            valign=center,
            nobeforeafter,
            boxrule=0.5pt
        ]
        \small Randomly switches from Norwegian to English. The Norwegian is correctly spelled but barely coherent.
        \end{tcolorbox}
    \end{minipage}
    \hfill
    \begin{minipage}{0.48\textwidth}
        \begin{tcolorbox}[
            title={llama3.3 70B:},
            colframe=boxborder,
            colback=boxbackground,
            coltitle=boxtitle,
            coltext=boxcontent,
            fonttitle=\bfseries,
            height=3.5cm
        ]
        \color{norwegian}
        Ifølge norsk historie og tradisjon var Harald Hårfagre Norges første konge. Han regjerte fra ca. 872 til
        933 og er kjent for å ha samlet Norge under ett styre etter slaget på Hafrsfjord i 872...
        \end{tcolorbox}
        \begin{tcolorbox}[
            colframe=boxborder,
            colback=notebg,
            height=1.7cm,
            valign=center,
            nobeforeafter,
            boxrule=0.5pt
        ]
        \small Maintains coherent Norwegian throughout the response. Provides factually correct historical context and dates.
        \end{tcolorbox}
    \end{minipage}
    \caption[]{A comparison of the first few sentences of an older and a newer iteration of Llama LLM prompted to answer in Norwegian. The Norwegian language writing capabilities of open models have improved dramatically over the last few years. Note that the differences presented here are exaggerated due to the selected examples, and even more so because of the difference in model size. While the newer generations of Llama are significantly better at speaking Norwegian, this is generally only evident after a longer conversation.}
    \label{fig:nor-conv}
\end{figure*}

\section{Related Work}
The concept of quantization is nothing new to the field of machine learning. According to Grey and Neuhoff's survey paper published in 1998, these techniques stem all the way back to 1948 with work relating to pulse-
code modulation systems. In this era, quantization generally refers to the mapping of continuous signals into discrete encoding\cite{Quantization}. In more recent times, quantization has gained new relevance as a method for reducing the size and computational cost of ML models, especially in the context of low-power computing. A relatively recent study of these methods has been published by Gholani et al. in 2022\cite{2022Study}.\\

A previous study by Marchisio et al. has investigated this topic for other quantization schemes and found the impact on multilingual performance to be present, but negligible. Across several benchmarks, the performance drop for multilingual tasks is generally measured to be lower than 3\%. This decline is more noticeable to human annotators, however, with an average performance drop of $13.1\%$ across all categories. Further, languages with a non-Latin script are generally more severely affected by quantization on automated benchmarks\cite{multilingualQuant}.\\

An LLM's performance in different languages varies significantly based on its prevalence within its pre-training corpus. Li et al. propose and provide a benchmarking of language performance in a recent paper\cite{languageRanker}.


\subsection{Previous investigation into Multilingual LLMs} \label{sec:prev-inv}
Some previous work exists investigating the inner workings of LLMs concerning multilingualism. Zhao et al. released a paper looking into the neurons of an LLM and their impact on its performance on multilingual tasks. They hypothesize that LLMs take multilingual input, translate it into English, process it, and then translate it back into the input language.
\begin{figure}
    \centering
    \includegraphics[width=0.75\linewidth]{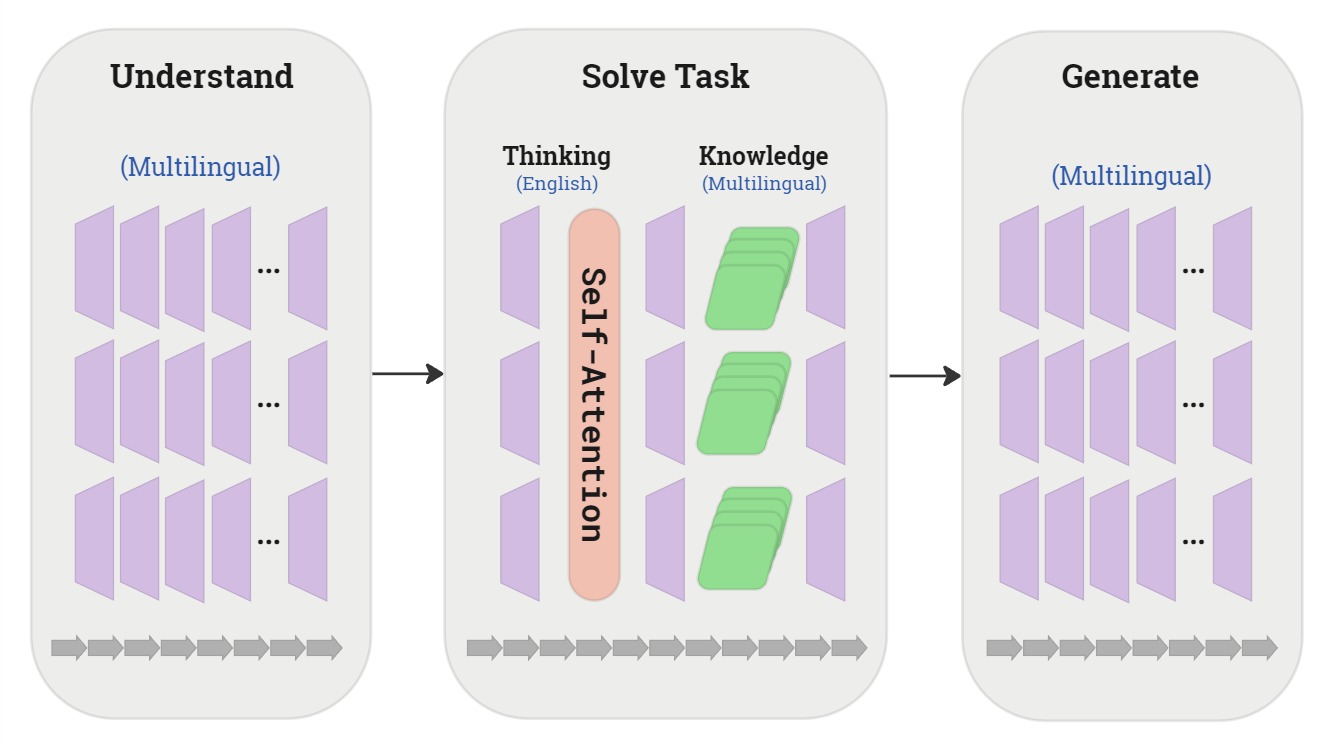}
    \caption{The hypothesized multilingual workflow of Zhao et al. In their view, LLMs convert multilingual queries to English for reasoning in English before it is converted back into the original language. Figure has been recreated based on one from the original publication of Zhao et al.\cite{multilingualLLM}}
    \label{fig:zhao-hypo}
\end{figure}
Through the application of their framework, they detected a set of neurons that were heavily tied to multilingual performance. By disabling the subset of 0.13\% of all neurons, performance on multilingual summarization tasks fell by 99\%.\cite{multilingualLLM}

\section{Method}
Few existing open-source LLMs explicitly support Norwegian; however, most attain some level of proficiency in the language passively. We will be using llama3.3 as our initial experimental model. This family of models supports eight languages (English, German, French, Italian, Portuguese, Hindi, Spanish, and Thai)\cite{llama3herd}. All experiments will be carried out on the relatively large llama3.3 70b LLMs due to their respectable performance on Norwegian language tasks. Based on our experience, LLMs require roughly this number of parameters to perform well in such a minor language without having been explicitly trained to support it.\\


Several different datasets exist that may be used to evaluate the performance of an LLM. One of the more established of these is MixEval, which was released in June 2024. The dataset notably has a 0.96 model rank correlation with the Chatbot Arena leaderboard\cite{ChatbotArena}— one of the more well-respected evaluation methods available today, without incurring the cost of human evaluation. As prominent datasets for LLM evaluation tend to contaminate training data, MixEval proposes a novel pipeline for dynamically building a dataset based on a mixture of existing LLM benchmarks. The dataset is updated regularly to maintain its integrity in the face of contamination. We will be using the $2024-06-01$ for our evaluation, a date well beyond the Llama3.3 data cut-off date of $2023-12-01$. The dataset is split into two portions, MixEval and MixEval-hard, in which the hard section represents questions that are deliberately picked to distinguish between stronger models. For this project, we will be using the standard difficulty version of the dataset\cite{MixEval}.\\

K-quantization and the application of an importance matrix are methods for quantization developed by Iwan Kawrakow. While these methods are not published in peer-reviewed form
, it has become ubiquitous due to its inclusion in the llamacpp library—as evidenced by how prevalent the quantization scheme is for the .gguf format\cite{kquant,imatrix}. While other approaches arrive at the importance matrix by looking at the weights gradient from a training run on a set of training tokens, this implementation calculates an approximation of these values based on the activations on a set of inferences.\\


\subsection{Importance Matrix}
One of the more prominent providers of LLM GGUF files is the Huggingface user Colin 'Bartowski' Kealty, who kindly provides the source text used for his importance matrices. The document consists primarily of English snippets of text from several diverse domains compiled with the help of users "Dampf" and "Kalomaze". This source text has been machine-translated via GPT-4o into Norwegian for our own quantization purposes, of which $0.63\%$ has been pruned due to being untranslatable. The pruned section primarily consisted of riddles built around spelling and wordplay\footnote{For example:\\Riddle: What is 3/7 chicken, 2/3 cat and 2/4 goat?\\ Answer: Chicago}. Further, for our baseline importance matrix in English, the non-English sections have been removed.




\subsection{MixEval Experiment Parameters}
All questions are provided in both Norwegian and English. The temperature of the generation is set to 0 to ensure deterministic output. Prompts are modeled after the prompts listed in MixEval's GitHub repo. The LLM is constrained to answer a valid choice index within the first token of the response. The limitation of a single token response significantly hampers the LLM's performance on certain tasks but results in a more objective evaluation of its ability to solve a task in a single forward pass due to the inherent stochasticity of token generation. Further, the constraint towards valid responses could favor the weaker quantized models by preventing them from failing to follow the original prompt.\\

 
\section{Results \& Discussion}
Our overall findings indicate that English quantization is more performant than language-dedicated quantized models. With the starting hypothesis that quantizing based on a non-English language significantly alters the performance on a benchmark, no p-value is calculated to be lower than $0.2373$ for the multiple choice dataset—a clear failure to disprove the null hypothesis. P-values were calculated by Monte Carlo simulation with the assumption that the probability found by the English version of each quant tier models the underlying binomial distribution of the success chance for each question.\\

One possible flaw in our method has been the fact that our importance matrices are machine-translated from English via LLMs, this could potentially bias the importance matrices towards language that appears 'close to English' in the estimation of an LLM. However, since a different, significantly larger LLM with a wholly different architecture was used to translate the importance matrices, this effect should be minimal.

\colorlet{boxborder}{brown!60!black}   
\colorlet{boxbackground}{orange!5}     
\colorlet{boxtitle}{white!90!black}    
\colorlet{boxcontent}{black}           
\colorlet{notebg}{orange!10}            
\colorlet{norwegian}{teal!70!black}    
\colorlet{english}{red!50!black}       

\begin{figure}
\begin{tcolorbox}[colback=boxbackground, colframe=boxborder]
\color{boxcontent}
\textbf{Question:} When was the last time ku won the championship?

\textbf{Answer:} 2022
\end{tcolorbox}
\caption{One of the example questions from MixEval's five-shot template. To omit both the original language and cultural context of the question renders it incomprehensible. "ku" here is meant to refer to the Kansas Jayhawks, an American basketball team from Kansas University. A Llama3.3 70B instance is able to answer the original question without issue but fails if it is translated to Norwegian due to the lack of cultural context.}
\label{fig:mixeval_question}
\end{figure}

The drop in performance on the Norwegian dataset may be partially explained by errors in translation. Directly translating questions into another language will occasionally lead to an unintended increase in difficulty. Such as if the original language provided some cultural context or accidental inclusion of homonyms in the target language. As an example, in Figure \ref{fig:mixeval_question}, the fact that the question is written in English is a vital clue to understanding what is being asked. \\
\begin{figure*}[h]
    \centering
    \includegraphics[width=1\linewidth]{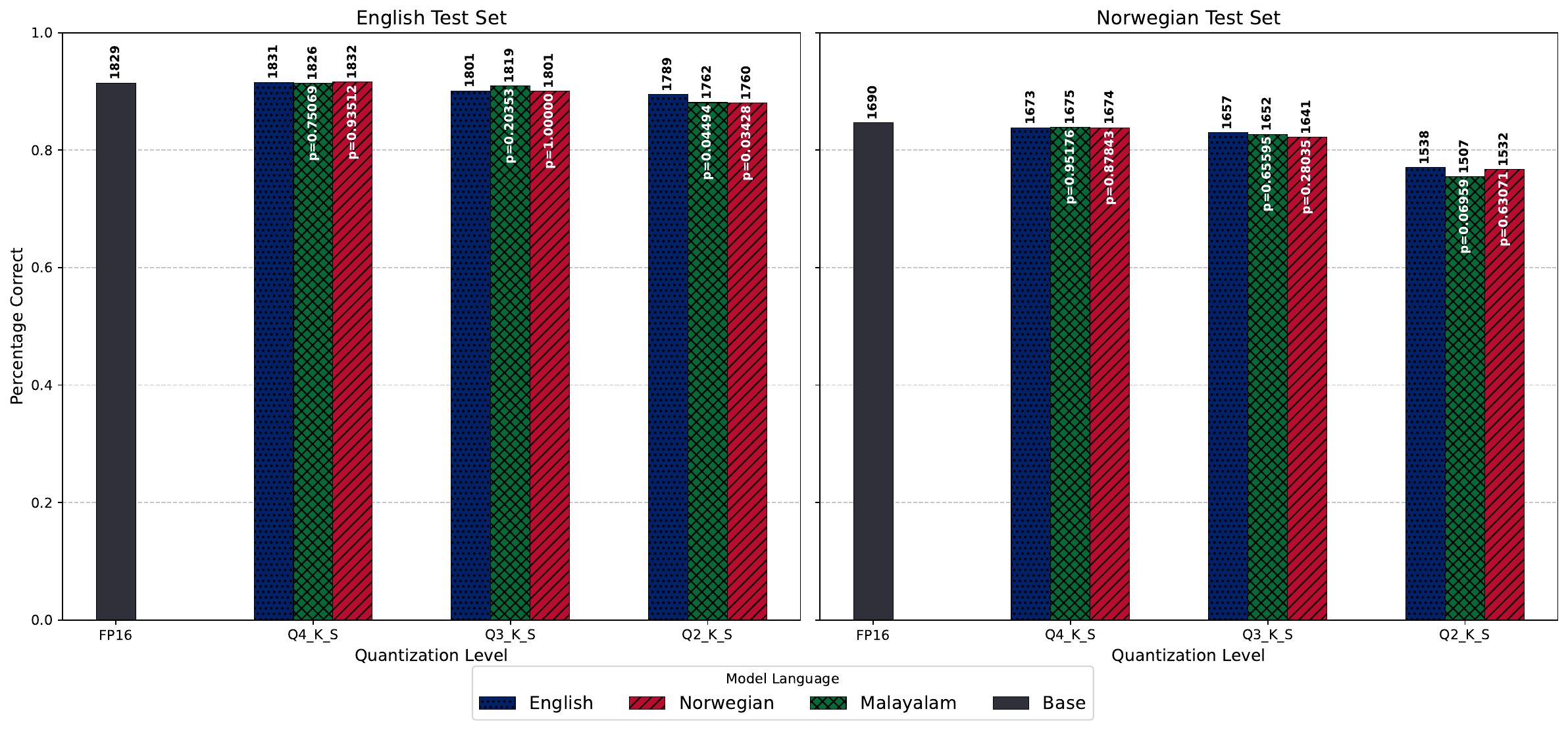}
    \caption{Comparison of percentage correct answers for three different language quants on both the English and Norwegian datasets for MixEval multiple choice questions. The total correct answers out of 2000 are displayed above each bar. P values for each of the results are displayed in white horizontally along the bar. Though quantized using a Norwegian importance matrix, our results indicate no statistically significant improvement on Norwegian MCQ when compared to the original English quantization, nor does the unrelated language of Malayalam significantly reduce performance in either language.}
    \label{fig:mixeval-mqc-results}
\end{figure*}

It seems unlikely to us that the pipeline as mapped out by Zhao et al. in Figure \ref{fig:zhao-hypo} explains the entirety of multilingual behavior within LLMs. Firstly, because machine learning at this scale rarely results in such tidy behavior, and secondly, if this were the case, its ability to respond to multilingual queries would be equal to its ability to translate the given language. Qualitatively, this is not the case; translation is a significantly simpler task than instruction following, in our experience. However, our results do not contradict their hypothesis as performance on both Norwegian and Malayalam tasks appears heavily connected to performance in English.\\

\begin{table}[htbp]
    \caption{Relative File Sizes (identical across eng, nor, and mal languages)}
    \centering
        \begin{tabular}{ccccc}
        \toprule
        & \textbf{fl16} & \textbf{Q4\_K\_5} & \textbf{Q3\_K\_5} & \textbf{Q2\_K\_5} \\
        \midrule
        \textbf{File Size} & 131.4 GB & 37.6 GB & 28.8 GB & 22.8 GB \\
        \textbf{Compression Rate} & $1.0\times$ & $\sim3.5\times$ & $\sim4.6\times$ & $\sim5.8\times$ \\
        \bottomrule
    \end{tabular}
    
    \vspace{0.5cm} 
    
    \label{tab:filesizes}
\end{table}

A graph of the results for the different quantization methods on the multiple choice questions is shown in Figure \ref{fig:mixeval-mqc-results}. There is a slight, but perceptible decline for each successive quant, though this loss is remarkably low when compared to their compression rate, as seen in Table \ref{tab:filesizes}. The accuracy of all models also noticeably degrades on the Norwegian dataset as compared to the original English version. This decline in performance is likely explained by a combination of poorer overall performance in Norwegian and an inability to translate technical terms into Norwegian. Ironically, it appears that some of the Norwegian quants perform worse on the Norwegian dataset as compared to the other language quants. However, based on how noisy the results are, the variance is likely due to random chance rather than any pattern between the quants.\\

\begin{figure*}[h]
    \centering
    \includegraphics[width=1\linewidth]{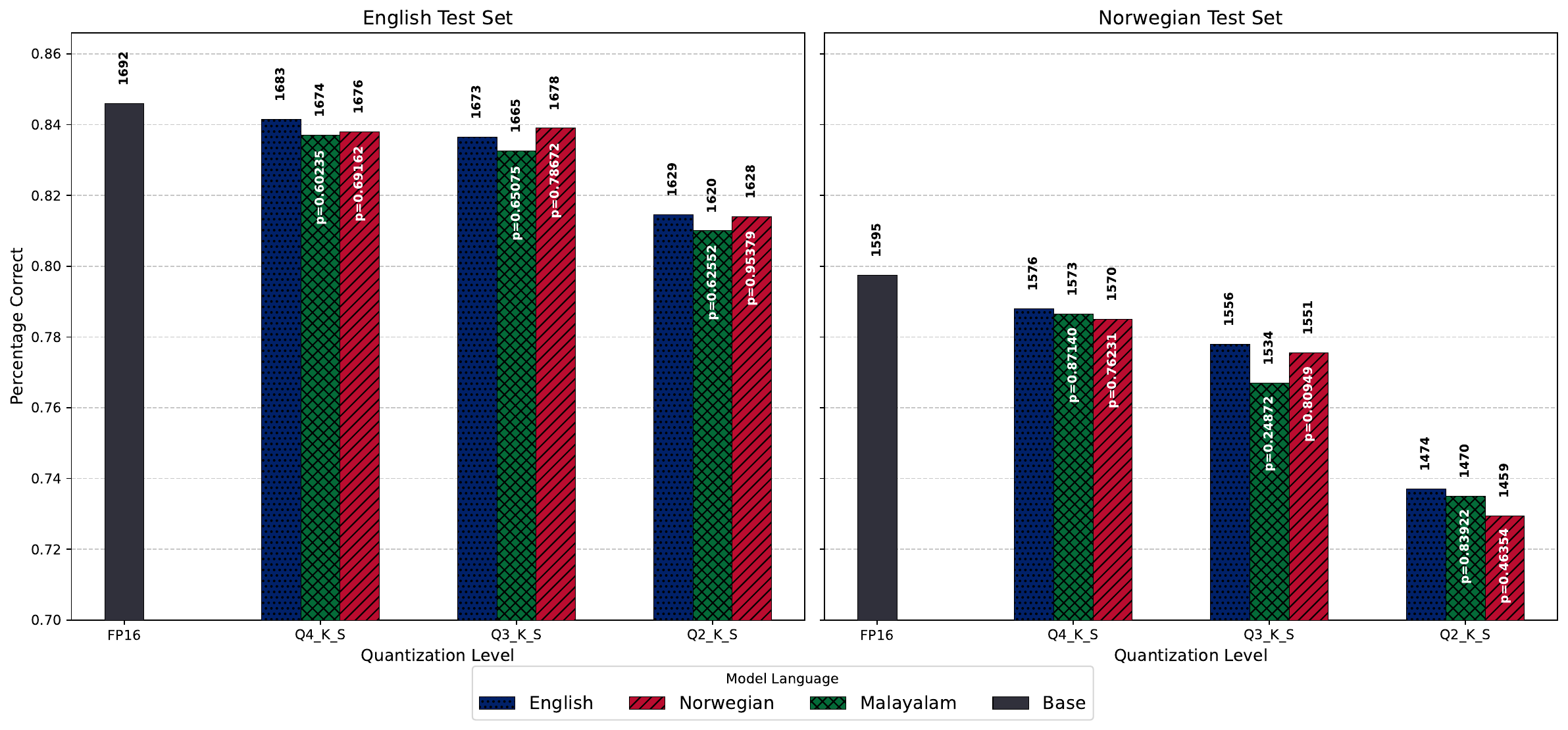}
    \caption{Comparison of performance percentage for three different language quants on both the English and Norwegian datasets of MixEval freeform questions. Total points out of a possible 2000 are displayed above each bar. P values for each of the results are displayed in white horizontally along the bar. To improve legibility, the y-axis has been truncated to the range from 0.7 to 0.9. Though quantized using a Norwegian importance matrix, our results indicate no statistically significant improvement on Norwegian free-form questions when compared to the original English quantization, nor does the unrelated language of Malayalam significantly reduce performance on either language.}
    \label{fig:mixeval-freeform-results}
\end{figure*}

The results for the free-form portion of MixEval can be seen in Figure \ref{fig:mixeval-freeform-results}. Similarly to the multiple choice questions, the accuracy of the different language quants performs similarly to the English version. Both the Norwegian and Malayalam quants of Q2\_K\_S arrive at p values lower than the standard significance threshold of 0.05 in English, given that they were evaluated in isolation. We should keep in mind that these experiments were performed with a common hypothesis for all of them, however, and therefore correct our significance threshold based on the number of experiments performed. By applying, for example, the Bonferroni correction\cite{bonferroni} and adjusting the threshold to $\alpha = \frac{original\ \alpha}{number of experiments}$. The resulting significance threshold of $\frac{0.05}{12} = 0.00417$ is still well below the lowest found p-value.

\subsection{Limitations}
Despite the number of experiments already performed, our investigations have only scratched the surface of the number of combinations of LLMs, languages, and quantization schemes to evaluate. A more thorough exploration would account for a wider range of these combinations; however, the results presented in this article provide an indication of how well other combinations of models and quantization schemes will perform.\\


Marchisio et al.\cite{multilingualQuant} noted a significant decrease in how natural multilingual speech appears as compared to English. This effect is difficult to quantify without human evaluation from native speakers. Given the resources to hire human annotators, as Marchisio et al. chose to do, it is still possible that our language-specific quants would outperform the English variants in this respect.

\section{Conclusion}
We have conducted a series of experiments quantizing Llama 3.3 70b instances on multilingual importance matrices. Our finding indicate that k\_quantization does not disproportionately affect multilingual performance. I.e., the application of English importance matrices does not appear to disproportionately reduce a model's scores on multilingual versions of the MixEval dataset. Further, the usage of importance matrices written in non-English does not significantly improve performance on non-English datasets and might in fact \emph{slightly} harm it. However, this reduction in performance is not found to be statistically significant.
%
%
%
%
\bibliographystyle{splncs04}
\bibliography{citations}

@article{Quantization,
	title = {Quantization},
	volume = {44},
	copyright = {https://ieeexplore.ieee.org/Xplorehelp/downloads/license-information/IEEE.html},
	issn = {00189448},
	url = {http://ieeexplore.ieee.org/document/720541/},
	doi = {10.1109/18.720541},
	number = {6},
	urldate = {2025-02-17},
	journal = {IEEE Transactions on Information Theory},
	author = {Gray, R.M. and Neuhoff, D.L.},
	month = oct,
	year = {1998},
	pages = {2325--2383},
}

@incollection{2022Study,
	address = {Boca Raton},
	edition = {1},
	title = {A {Survey} of {Quantization} {Methods} for {Efficient} {Neural} {Network} {Inference}},
	isbn = {978-1-003-16281-0},
	url = {https://www.taylorfrancis.com/books/9781003162810/chapters/10.1201/9781003162810-13},
	language = {en},
	urldate = {2025-02-17},
	booktitle = {Low-{Power} {Computer} {Vision}},
	publisher = {Chapman and Hall/CRC},
	author = {Gholami, Amir and Kim, Sehoon and Dong, Zhen and Yao, Zhewei and Mahoney, Michael W. and Keutzer, Kurt},
	collaborator = {Thiruvathukal, George K. and Lu, Yung-Hsiang and Kim, Jaeyoun and Chen, Yiran and Chen, Bo},
	month = jan,
	year = {2022},
	doi = {10.1201/9781003162810-13},
	pages = {291--326},
}

@misc{kquant,
  author       = {Iwan 'ikawrakow' Kawrakow},
  title        = {k-quants},
  howpublished = {\url{https://github.com/ggml-org/llama.cpp/pull/1684}},
  year         = {2023},
  note         = {Accessed: 2025-02-18}
}

@misc{imatrix,
  author       = {Iwan 'ikawrakow' Kawrakow},
  title        = {Importance Matrix calculation},
  howpublished = {\url{https://github.com/ggml-org/llama.cpp/pull/4861}},
  year         = {2023},
  note         = {Accessed: 2025-02-18}
}

@misc{llama3herd,
	title = {The {Llama} 3 {Herd} of {Models}},
	url = {http://arxiv.org/abs/2407.21783},
	doi = {10.48550/arXiv.2407.21783},
	urldate = {2025-02-19},
	publisher = {arXiv},
	author = {Grattafiori, Aaron and et al},
	month = nov,
	year = {2024},
	note = {arXiv:2407.21783 [cs]},
}

@misc{MixEval,
	title = {{MixEval}: {Deriving} {Wisdom} of the {Crowd} from {LLM} {Benchmark} {Mixtures}},
	shorttitle = {{MixEval}},
	url = {http://arxiv.org/abs/2406.06565},
	doi = {10.48550/arXiv.2406.06565},
	urldate = {2025-02-19},
	publisher = {arXiv},
	author = {Ni, Jinjie and Xue, Fuzhao and Yue, Xiang and Deng, Yuntian and Shah, Mahir and Jain, Kabir and Neubig, Graham and You, Yang},
	month = oct,
	year = {2024},
	note = {arXiv:2406.06565 [cs]},
}

@misc{ChatbotArena,
	title = {Chatbot {Arena}: {An} {Open} {Platform} for {Evaluating} {LLMs} by {Human} {Preference}},
	shorttitle = {Chatbot {Arena}},
	url = {http://arxiv.org/abs/2403.04132},
	doi = {10.48550/arXiv.2403.04132},
	urldate = {2025-02-19},
	publisher = {arXiv},
	author = {Chiang, Wei-Lin and Zheng, Lianmin and Sheng, Ying and Angelopoulos, Anastasios Nikolas and Li, Tianle and Li, Dacheng and Zhang, Hao and Zhu, Banghua and Jordan, Michael and Gonzalez, Joseph E. and Stoica, Ion},
	month = mar,
	year = {2024},
	note = {arXiv:2403.04132 [cs]},
}

@misc{ethnologue200,
  author       = {Eberhard, David M. and Simons, Gary F. and Fennig, Charles D.},
  title        = {Ethnologue 200: What are the top 200 most spoken languages?},
  year         = {2024},
  howpublished = {\url{https://www.ethnologue.com/insights/ethnologue200/}},
  note         = {Accessed: 2025-02-20},
}

@misc{deepseekV3,
	title = {{DeepSeek}-{V3} {Technical} {Report}},
	url = {http://arxiv.org/abs/2412.19437},
	doi = {10.48550/arXiv.2412.19437},
	urldate = {2025-02-20},
	publisher = {arXiv},
	author = {Liu, Aixin and et al.},
	month = feb,
	year = {2025},
	note = {arXiv:2412.19437 [cs]},
}

@misc{multilingualLLM,
	title = {How do {Large} {Language} {Models} {Handle} {Multilingualism}?},
	url = {http://arxiv.org/abs/2402.18815},
	doi = {10.48550/arXiv.2402.18815},
	urldate = {2025-02-20},
	publisher = {arXiv},
	author = {Zhao, Yiran and Zhang, Wenxuan and Chen, Guizhen and Kawaguchi, Kenji and Bing, Lidong},
	month = nov,
	year = {2024},
	note = {arXiv:2402.18815 [cs]},
}

@misc{languageRanker,
	title = {Language {Ranker}: {A} {Metric} for {Quantifying} {LLM} {Performance} {Across} {High} and {Low}-{Resource} {Languages}},
	shorttitle = {Language {Ranker}},
	url = {http://arxiv.org/abs/2404.11553},
	doi = {10.48550/arXiv.2404.11553},
	urldate = {2025-02-24},
	publisher = {arXiv},
	author = {Li, Zihao and Shi, Yucheng and Liu, Zirui and Yang, Fan and Payani, Ali and Liu, Ninghao and Du, Mengnan},
	month = dec,
	year = {2024},
	note = {arXiv:2404.11553 [cs]},
}

@misc{llama2,
	title = {Llama 2: {Open} {Foundation} and {Fine}-{Tuned} {Chat} {Models}},
	shorttitle = {Llama 2},
	url = {http://arxiv.org/abs/2307.09288},
	doi = {10.48550/arXiv.2307.09288},
	urldate = {2025-02-24},
	publisher = {arXiv},
	author = {Touvron, Hugo and Martin, Louis and Stone, Kevin and Albert, Peter and Almahairi, Amjad and Babaei, Yasmine and Bashlykov, Nikolay and Batra, Soumya and Bhargava, Prajjwal and Bhosale, Shruti and Bikel, Dan and Blecher, Lukas and Ferrer, Cristian Canton and Chen, Moya and Cucurull, Guillem and Esiobu, David and Fernandes, Jude and Fu, Jeremy and Fu, Wenyin and Fuller, Brian and Gao, Cynthia and Goswami, Vedanuj and Goyal, Naman and Hartshorn, Anthony and Hosseini, Saghar and Hou, Rui and Inan, Hakan and Kardas, Marcin and Kerkez, Viktor and Khabsa, Madian and Kloumann, Isabel and Korenev, Artem and Koura, Punit Singh and Lachaux, Marie-Anne and Lavril, Thibaut and Lee, Jenya and Liskovich, Diana and Lu, Yinghai and Mao, Yuning and Martinet, Xavier and Mihaylov, Todor and Mishra, Pushkar and Molybog, Igor and Nie, Yixin and Poulton, Andrew and Reizenstein, Jeremy and Rungta, Rashi and Saladi, Kalyan and Schelten, Alan and Silva, Ruan and Smith, Eric Michael and Subramanian, Ranjan and Tan, Xiaoqing Ellen and Tang, Binh and Taylor, Ross and Williams, Adina and Kuan, Jian Xiang and Xu, Puxin and Yan, Zheng and Zarov, Iliyan and Zhang, Yuchen and Fan, Angela and Kambadur, Melanie and Narang, Sharan and Rodriguez, Aurelien and Stojnic, Robert and Edunov, Sergey and Scialom, Thomas},
	month = jul,
	year = {2023},
	note = {arXiv:2307.09288 [cs]},
}

@misc{multilingualQuant,
	title = {How {Does} {Quantization} {Affect} {Multilingual} {LLMs}?},
	url = {http://arxiv.org/abs/2407.03211},
	doi = {10.48550/arXiv.2407.03211},
	urldate = {2025-02-27},
	publisher = {arXiv},
	author = {Marchisio, Kelly and Dash, Saurabh and Chen, Hongyu and Aumiller, Dennis and Üstün, Ahmet and Hooker, Sara and Ruder, Sebastian},
	month = oct,
	year = {2024},
	note = {arXiv:2407.03211 [cs]},
}

@article{Bonferroni,
	title = {Multiple comparison procedures},
	volume = {312},
	issn = {0098-7484},
	url = {https://doi.org/10.1001/jama.2014.9440},
	doi = {10.1001/jama.2014.9440},
	number = {5},
	journal = {JAMA : the journal of the American Medical Association},
	author = {Cao, Jing and Zhang, Song},
	month = aug,
	year = {2014},
	note = {tex.eprint: https://jamanetwork.com/journals/jama/articlepdf/1892228/jgm140005.pdf},
	pages = {543--544},
}

@misc{huggingfacemodelsno,
  author       = {Hugging Face},
  title        = {Models - Hugging Face},
  howpublished = {\url{https://huggingface.co/models?language=no,nn}},
  year         = {2025},
  note         = {Accessed: 2025-09-04}
}

@misc{nbllama,
  author       = {National Library of Norway (NB-AiLab)},
  title        = {NbAiLab/nb-llama-3.2-1B},
  howpublished = {\url{https://huggingface.co/NbAiLab/nb-llama-3.2-1B}},
  year         = {2025},
  note         = {Accessed: 2025-09-04}
}

@misc{nbllamapaper,
	title = {Operationalizing a {National} {Digital} {Library}: {The} {Case} for a {Norwegian} {Transformer} {Model}},
	shorttitle = {Operationalizing a {National} {Digital} {Library}},
	url = {http://arxiv.org/abs/2104.09617},
	doi = {10.48550/arXiv.2104.09617},
	urldate = {2025-09-05},
	publisher = {arXiv},
	author = {Kummervold, Per E. and Rosa, Javier de la and Wetjen, Freddy and Brygfjeld, Svein Arne},
	month = apr,
	year = {2021},
	note = {arXiv:2104.09617 [cs]},
}

@inproceedings{norBERT,
    title = "Large-Scale Contextualised Language Modelling for {N}orwegian",
    author = "Kutuzov, Andrey  and
      Barnes, Jeremy  and
      Velldal, Erik  and
      {\O}vrelid, Lilja  and
      Oepen, Stephan",
    editor = "Dobnik, Simon  and
      {\O}vrelid, Lilja",
    booktitle = "Proceedings of the 23rd Nordic Conference on Computational Linguistics (NoDaLiDa)",
    month = may # " 31--2 " # jun,
    year = "2021",
    address = "Reykjavik, Iceland (Online)",
    publisher = {Link{\"o}ping University Electronic Press, Sweden},
    url = "https://aclanthology.org/2021.nodalida-main.4/",
    pages = "30--40"
}

\end{document}